\documentclass{article}

\usepackage{PRIMEarxiv}

\usepackage[utf8]{inputenc} 
\usepackage[T1]{fontenc}    
\usepackage{url}            
\usepackage{booktabs}       
\usepackage{amsfonts}       
\usepackage{nicefrac}       
\usepackage{microtype}      
\usepackage{lipsum}
\usepackage{fancyhdr}       
\usepackage{graphicx}       
\graphicspath{{media/}}     

\title{From Text to Motion: Grounding GPT-4\\ in a Humanoid Robot ``Alter3''}

\author{Takahide Yoshida$^{1,*}$, Atsushi Masumori $^{1,2, \dag}$ and  Takashi Ikegami$^{1,2,\ddag}$ \\
\mbox{}\\
\date{}
$^1$Graduate School of Arts and Sciences, The University of Tokyo, Tokyo, Japan \\
$^2$Alternative Machine Inc., Tokyo, Japan \\
\{$^*$yoshida, $^\dag$masumori, $^\ddag$ikeg\}@sacral.c.u-tokyo.ac.jp \\} 

\usepackage{xcolor} 
\usepackage{pdfpages}
\usepackage{amsmath}
\usepackage{tcolorbox}
\tcbset{
  colback=gray!5,       
  colframe=gray!50,     
  fonttitle=\bfseries,  
  coltitle=black,       
  center title,         
  rounded corners,      
  fontupper=\ttfamily,  
}

\usepackage{listings}
\lstdefinestyle{mystyle}{
    language=Python,
    backgroundcolor=\color{gray!10}, 
    commentstyle=\color{green},
    keywordstyle=\color{blue},
    numberstyle=\tiny\color{gray},
    stringstyle=\color{red},
    basicstyle=\ttfamily\footnotesize,
    breakatwhitespace=false,
    breaklines=true,
    captionpos=b,
    keepspaces=true,
    numbers=left,
    numbersep=5pt,
    showspaces=false,
    showstringspaces=false,
    showtabs=false,
    tabsize=2
}
\usepackage[colorlinks=true, allcolors=blue]{hyperref}

\usepackage{enumitem}
\setlist{nosep}

\begin{document}
\maketitle

\begin{abstract}

We report the development of Alter3, a humanoid robot capable of generating spontaneous motion using a Large Language Model (LLM), specifically GPT-4. This achievement was realized by integrating GPT-4 into our proprietary android, Alter3, thereby effectively grounding the LLM with Alter's bodily movement. Typically, low-level robot control is hardware-dependent and falls outside the scope of LLM corpora, presenting challenges for direct LLM-based robot control. However, in the case of humanoid robots like Alter3, direct control is feasible by mapping the linguistic expressions of human actions onto the robot's body through program code. Remarkably, this approach enables Alter3 to adopt various poses, such as a 'selfie' stance or 'pretending to be a ghost,' and generate sequences of actions over time without explicit programming for each body part. This demonstrates the robot's zero-shot learning capabilities. Additionally, verbal feedback can adjust poses, obviating the need for fine-tuning. A video of Alter3's generated motions is available at \href{https://tnoinkwms.github.io/ALTER-LLM/}{this URL}.

\end{abstract}

\section{Introduction}
Recent advancements in artificial intelligence have seen the development of large language models (LLMs), which demonstrate an impressive ability to generate contextually relevant text and code in response to various prompts. In particular, OpenAI's GPT-4 has remarkable inferential capabilities and a vast amount of knowledge through extensive text information\cite{openai2023gpt4}. Despite these capabilities, which enable LLMs to exhibit ``human-like'' behaviors in their interactions, they are not grounded in the physical world, and their expressions and activities are confined to the realm of text. We might wonder whether the vast knowledge that is encoded in LLMs can be used by humanoid robots. In our laboratory, we connect a humanoid robot ``Alter3'' with a LLM. This integration is aimed at opening the unforeseen potential of LLM. 


In the past few years, the integration of LLMs with robots has emerged as a promising frontier in the field of artificial intelligence and robotics. The applications of LLM in robotics are wide-ranging, such as human-robot interaction\cite{sun2023humanoid, zhang2023large}, task planning\cite{ding2023task, yu2023language}, navigation\cite{zeng2023socratic, huang2023visual}, learning \cite{shafiullah2023clipfields, zhong2023chatabl} and so on. Interdisciplinary approaches are particularly important for enhancing the cognitive and interactive capabilities of humanoid robots. Leveraging the prowess of LLMs in natural language processing is a key aspect of this advancement\cite{saycan, brohan2023rt2, liang2023code, driess2023palme}. In this domain of using LLM in robots, there have been primarily focused on facilitating basic communication between life and robots within a computer, utilizing LLMs to interpret and pretend life-like responses \cite{ochiai2023digital, liang2023code}. Moreover, researchers have begun exploring the potential of LLMs in enabling robots to understand and process complex instructions, thereby augmenting their functionality and autonomy.

For example, Ahn et al.'s robot called ``SayCan'' \cite{saycan} can extract from our speech and context to generate an essential message by using LLM and selects the most adequate action sequence that SayCan himself follows. However, low-level robot control is dependent on hardware, which is not included in the LLMs corpora, making it challenging for LLMs to directly control robots. Therefore, studies like Say-can require such as TD-learning (Time difference) and reward functions for robot control\cite{saycan, yu2023language, tang2023saytap}. In this paper, we say that ``LLM knows how to map between human posture and language expressions.'' It is expected that human motion descriptions are more common than those for robotic arms or quadrupeds. Consequently, we suggest that humanoid robots, resembling humans in shape, could achieve precise movements through zero-shot learning, as explained in this paper.

Recent advancements are steering towards a more nuanced collaboration between LLMs and robots. A notable trajectory is the development of empathy and socially-aware robots that can perceive and adapt to human emotions and social cues, fostering a more natural and engaging human-robot interaction (\cite{ALTER, disney, ye2023improved}, \href{https://www.engineeredarts.co.uk/}{Ameca by Engineered Arts}). The incorporation of LLMs should be explored to enhance the learning capabilities of robots, allowing them to acquire new skills and knowledge through natural language instructions, a paradigm shift from traditional programming-based approaches. This is what this article also tries to focus on. 

We have been developing a new humanoid robot since 2016 called ``Alter series''\cite{idoi}  and the third version of Alter (Alter3 in short) is used in the current study to produce new behaviors with LLM (the architecture of Alter3 will be described in the next section). As this kind of experiment (Robot + LLM) continues to evolve, it holds the potential to redefine the boundaries of human-robot collaboration, paving the way for more intelligent, adaptable, and personable robotic entities. Embodied LLMs represent a groundbreaking paradigm in the study of human psychology, cognitive science, and the philosophy of embodied cognition, as highlighted at CogSci2023 in Sydney (July 23-26, 2023). This paper demonstrates how motion can be generated from language using LLMs. Additionally, we develop a conversation system where LLM agents communicate with humans, incorporating physical gestures. By integrating LLMs with humanoid robots, we delve into the novel potentials of LLM technology.

\section{Design and Features of Alter3}
\begin{figure}
\centering
\includegraphics[width=0.6\linewidth]{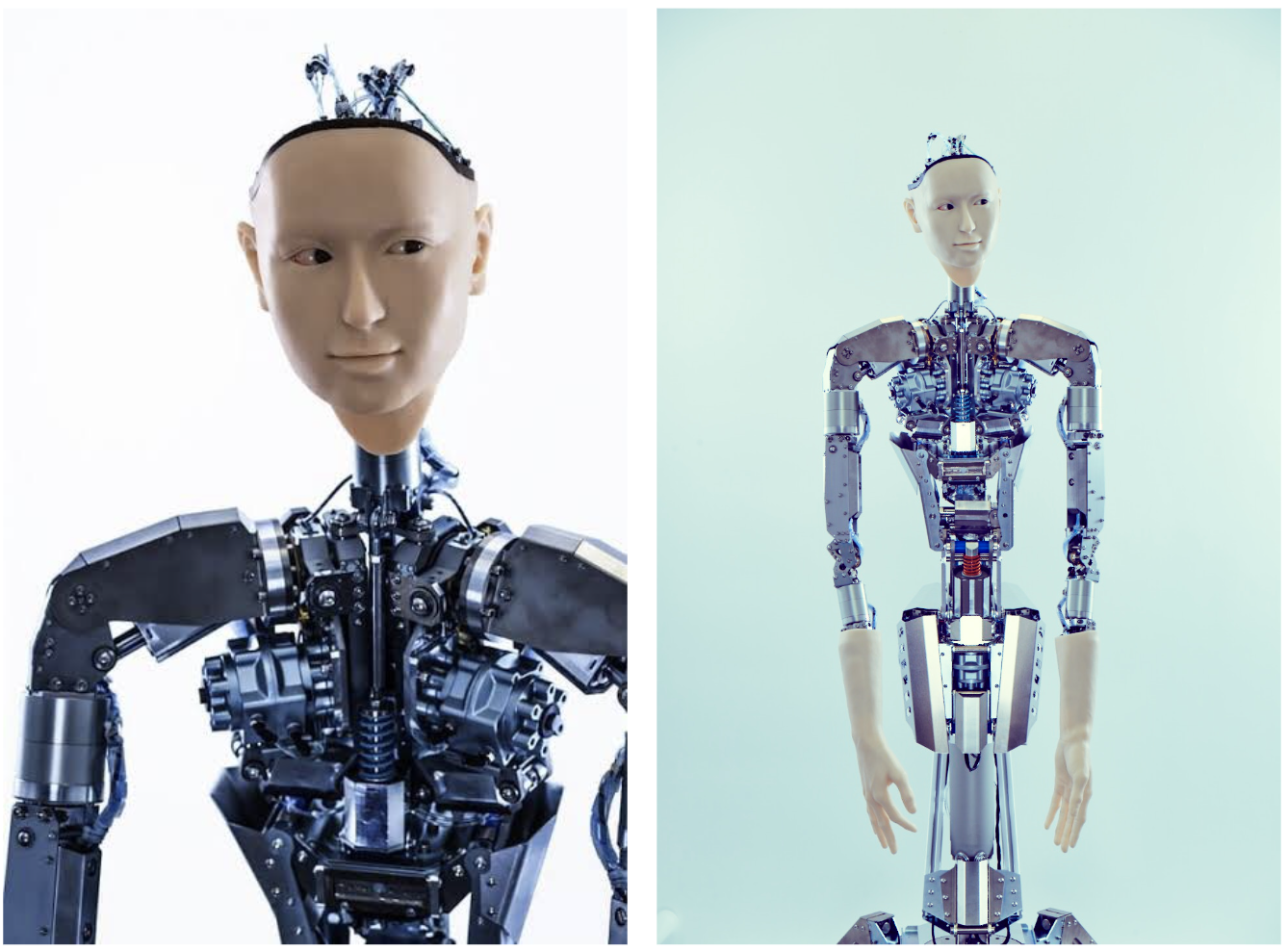}
\caption{\label{fig:alter3}Body of Alter3. The body has 43 axes that are controlled by air actuators. It is equipped with a camera inside each eye. The control system sends commands via a serial port to control the body. The refresh rate is 100–150 ms.}
\end{figure}


Alter3 represents the third iteration in the Alter humanoid robot series since 2016 (see Figure. \ref{fig:alter3}). It is equipped with 43 actuators, including facial expressions and limb movements, powered by compressed air, which enable a wide range of expressive gestures. However, Alter3 does not possess the ability to walk, although it can pretend walking and running motions (as we see below).

In our previous research, Alter3 demonstrated the ability to imitate human postures using a camera and the OpenPose framework\cite{openpose}. This imitation process involves adjusting its joints to replicate the observed poses and storing successful imitations for future reference. The effectiveness of these imitations is measured by the increase in transfer entropy, indicating information flow from humans to the robot\cite{ALTER,vae}. Further experiments involved mutual imitation scenarios with humans and between Alter2 and Alter3, revealing that interaction with humans leads to more varied poses\cite{Coupled_ALTER}. This supports the idea that diverse movements are derived from human imitation, echoing the development of ``neonatal imitation'' observed in newborn babies\cite{newborn}.

These findings suggest a potential offloading mind concept and pave the way for investigating the imitative abilities of robots with extensive learned data, which Alter3 exemplifies with its integration of advanced LLMs.

\section{Generating humanoid motions from text}

\subsection{Prompt engineering for motion generation of humanoid}
\begin{figure}[ht]
\centering
\includegraphics[width=0.8\linewidth]{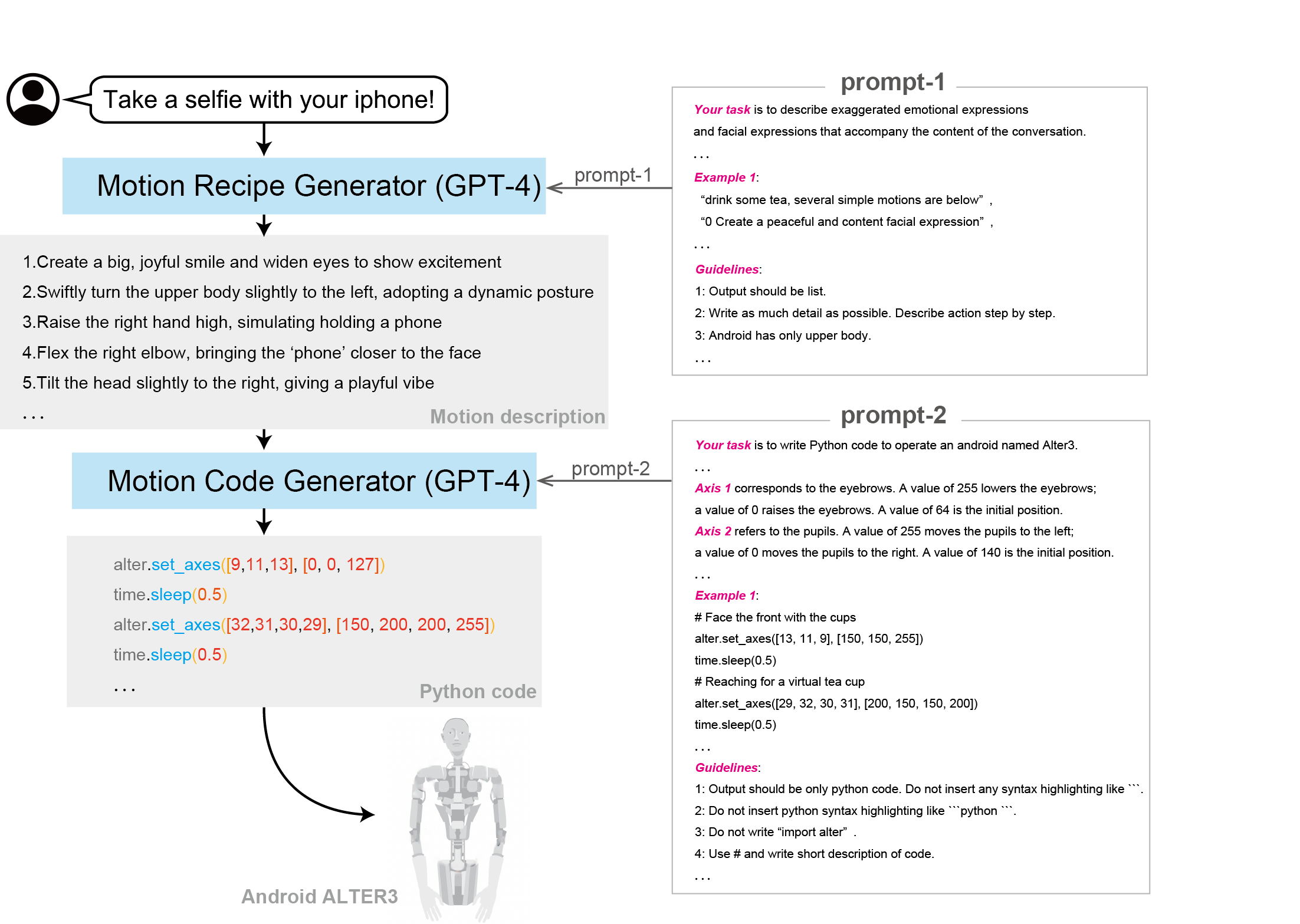}
\caption{\label{fig:prompt} A procedure to control the Alter3 humanoid  using verbal instructions. Output Python code to control Alter3 from natural language using prompt1 via prompt2. A humanoid robot, which mimics the human shape, can generate highly precise movements using zero-shot learning, eliminating the need for setting reward functions or interfaces, as required in other studies. The architecture is based on CoT. See the Appendix for details of the prompt.}
\end{figure}

Before the LLM appeared, we had to control all 43 axes in a certain order to mimic a person's pose or to pretend a behavior such as serving a tea or playing chess. The process usually required many refinements manually by ourselves. Thanks to LLM, we are now free from the iterative labor. 
Here is a procedure to control the Alter3 humanoid using verbal instructions.  What we do is to successively apply two protocols written in a natural language known as a chain of thought (CoT) \cite{wei2023chainofthought} and no iteration of a learning process is required (i.e. a zero shot learning). Practically speaking, we have used the following protocols\footnote{One important thing to note is that it’s well-known at this point that GPT-4 is non-deterministic, even at $temperature=0.0$. Therefore, even with identical inputs, different patterns of motion can be generated. This is a characteristic of OpenAI's GPT-4 and, while it poses an issue in terms of reproducibility, it should not be considered a reason to doubt its ability to generate movement.}. 

\noindent \textbf{Specify the action you want Alter3 to make.} e.g., Let's take a selfie with your iPhone!  \\
\textbf{prompt1:} LLM generates in about 10 lines with exaggerated descriptions of a given movement(see Figure \ref{fig:prompt}). An excerpt of the prompt is as follows. 
\begin{tcolorbox}
Your task is to describe exaggerated emotional expressions and facial expressions that accompany the content of the conversation. Output motion description should be several simple motions that the android is capable of. In addition, please create a facial expression that matches the input at the beginning. The android can only move its upper body and has the same joints as a human. Output should be written in as much detail as possible.
\end{tcolorbox}
\noindent We also provided an example of descriptions (which is also provided by LLM) and guidelines. These inputs and the own outputs are bases for creating an action pattern. By the next prompt, Alter3 will generate a Python code for organizing the action (see Appendix-prompt1).\\
\noindent \textbf{prompt2:} LLM creates python code from the output of prompt2. The meaning of Axis 1 to 43 are imported first. For example,
\begin{tcolorbox}
    Axis 1: Eyebrows. 255 = angry, 0 = surprised, 64 = neutral.\\
    Axis 2: Pupils (horizontal). 255 = left, 0 = right, 140 = neutral.\\
    Axis 3: Pupils (vertical). 255 = up, 0 = down, 128 = neutral.
\end{tcolorbox}
\noindent and so on. The descriptions of the motion direction such as 255 = angry or 0 = right are given by us. Again we provided an example of the Python code corresponding to ``drink some tea'' and guidelines.The output Python code will be automatically given to an android engine written in Python code to actually control the amplitude of the air compressor(see Appendix-prompt2). 

\subsubsection{Examples of generated actions}
\begin{figure}[ht]
\centering
\includegraphics[width=0.6\linewidth]{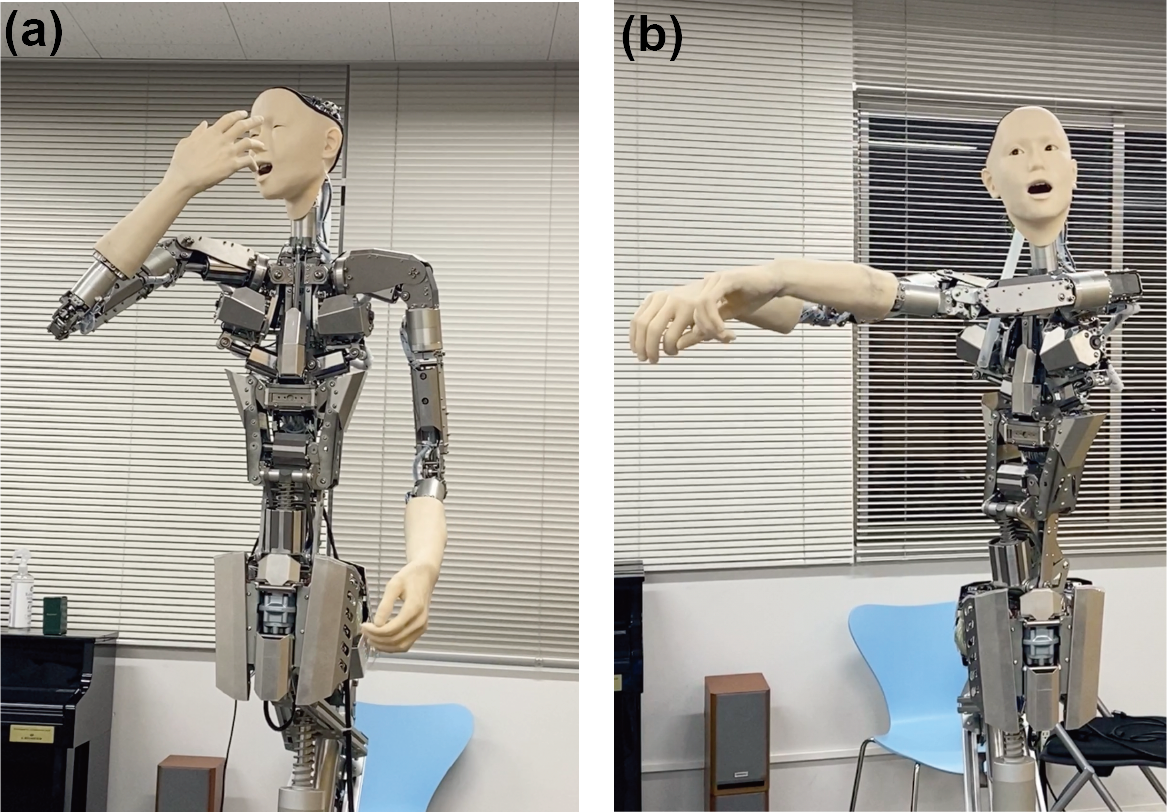}
\caption{\label{fig:snap} A snapshot of generated stereotypical movements. \textbf{(a)} take a selfie ($score=3.0$). \textbf{(b)} pretend a ghost ($score=3.1$). LLM can generate emotional expressions associated with specific movements. For example, in the case of a selfie, Alter3 is showing a smile.}
\end{figure}

\begin{figure}[ht]
\centering
\includegraphics[width=0.9\linewidth]{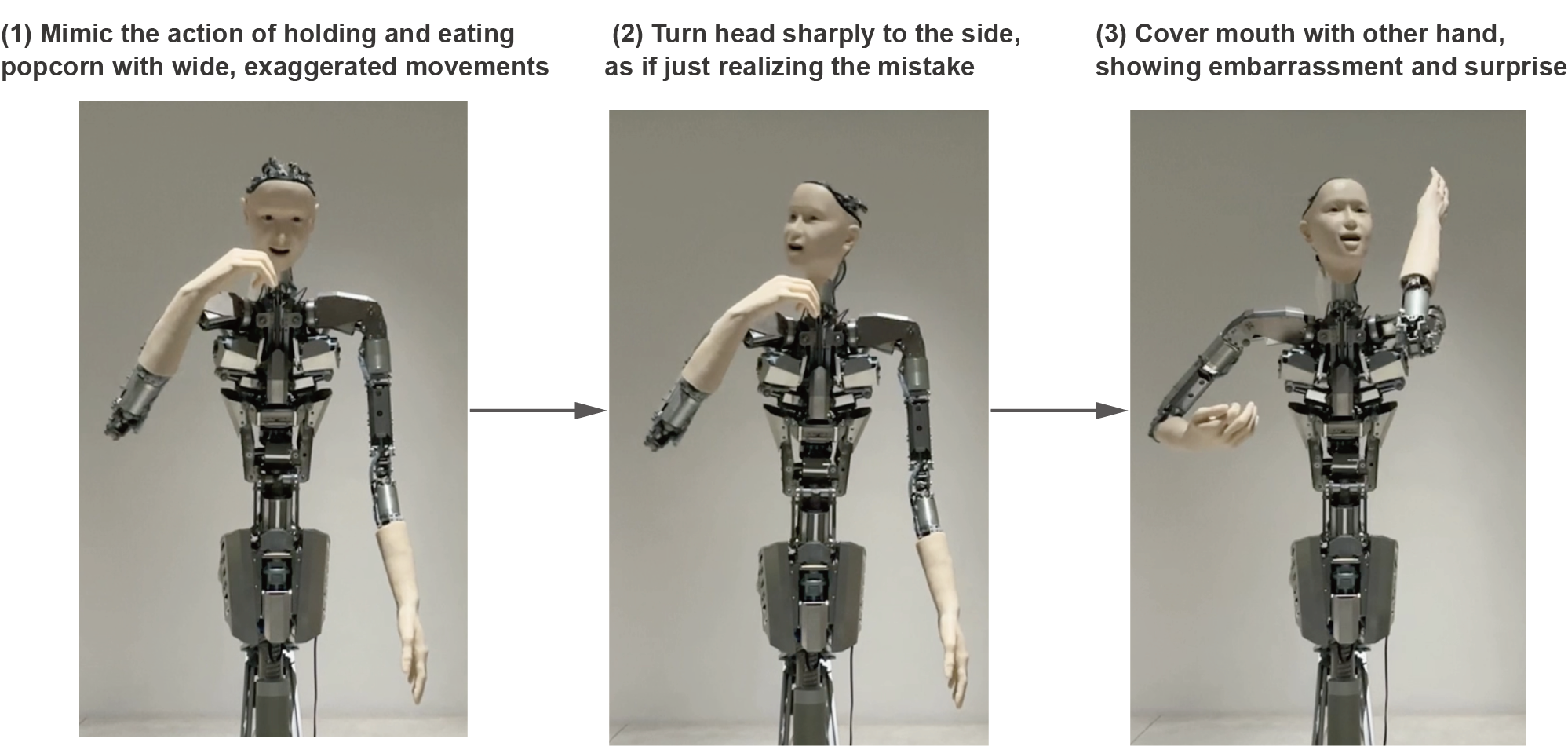}
\caption{\label{fig:popcorn} A snapshot of a generated sequence of movements: ``I was enjoying a movie while eating popcorn at the theater when I realized that I was actually eating the popcorn of the person next to me'' ($score=3.0$). LLM can generate movements that progress over time like a story. \textbf{Left}: The action of eating popcorn. \textbf{Center}: Noticing the person next to Alter3. \textbf{Right}: getting panicked.}
\end{figure}

Using the procedures, we have tested many actions and gestures taken by Alter3, such as "taking a selfie," "pretending to be a ghost," "throwing a ball," "playing the guitar," and the reactions of Alter3 when listening to a short story. Again, the example below demonstrates zero-shot learning; that is, we did not conduct any training or tuning except for the electric guitar example. Hence, the LLM possesses detailed knowledge of human movements, which can be executed in Alter3 via Python code. In the case of the electric guitar, we refined the code by verbally advising Alter3 (the feedback system will be explained in the next section). A video of Alter3’s generated motions is available at \href{https://tnoinkwms.github.io/ALTER-LLM/}{this url}.


The android Alter3 emulates motions and gestures that we perform consciously or unconsciously in daily life. We categorize these motions and gestures with respect to their time span:
\begin{enumerate}[label=\roman*)]
\item An instant action or gesture, e.g., taking an iPhone, pretending to be a ghost.
\item A gesture or action with a temporal sequence of events.
\end{enumerate}

We analyze how Alter3 expresses emotions through stereotypical movements. This involves evaluating its ability to generate appropriate responses to various statements. For instance, when generating actions like "taking a selfie" or "pretending to be a ghost" through GPT-4, Alter3 demonstrates actions often associated with emotional expressions.
A selfie action produces a smiling, joyful expression while pretending to be a ghost is a portrayal common among ordinary people. We automatically empathize with a person we are conversing with. When the other person tells a sad story, you also feel sad; when they tell a joke, you feel amused (see Figure \ref{fig:snap}). The second example is very different from the first, as the given description is declarative and sometimes abstract. Sometimes, there is no right gesture or action that Alter3 should take. Through the second example, we evaluated not only Alter3's instantaneous reaction capabilities but also its proficiency in emulating interconnected, human-like emotional expressions and movements over time.

For example, consider a scenario described as, "I was enjoying a movie while eating popcorn in the theater, when I suddenly realized that I was actually eating the popcorn of the person next to me." In response to this story, Alter3 would initially depict the act of eating popcorn and then transition to an expression of surprise upon realizing the mistake. This scenario demonstrates how actions and emotional responses can be sequenced to portray a realistic and continuous narrative, rather than just a single, instantaneous gesture (see Figure \ref{fig:popcorn}).

\subsection{Performance evaluation of LLM motion generation}
\begin{figure}[ht]
\centering
\includegraphics[width=1.0\linewidth]{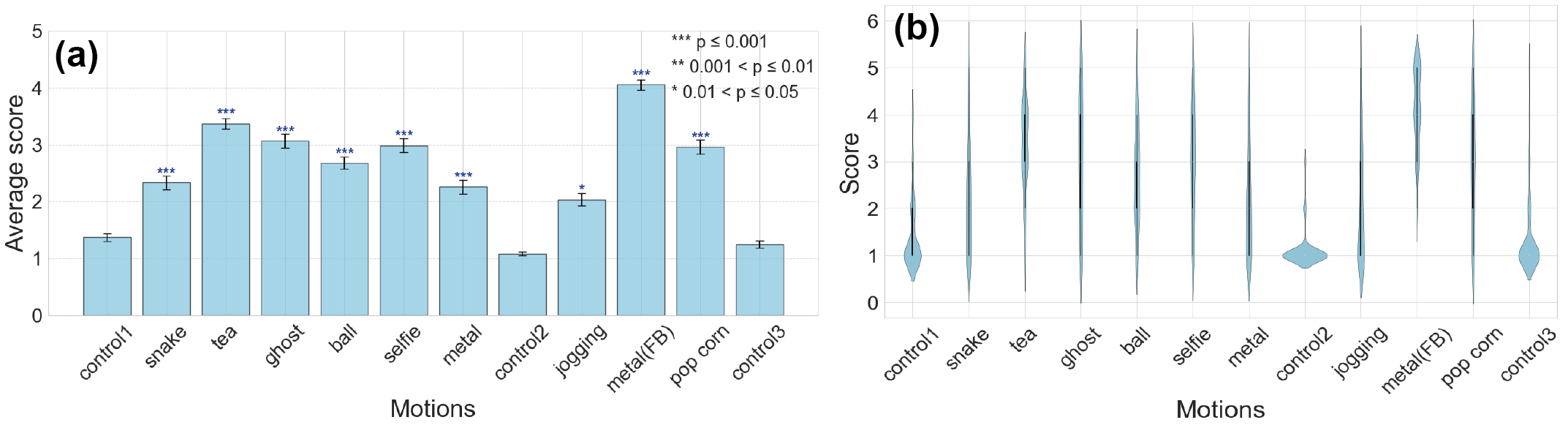}
\caption{\label{fig:score}Third-party evaluation of the generated motions. The following behaviors of Alter3 are evaluated by the subjects ($n=107$) recruited using platform Prolific; ``pretend the snake'', ``drink some tea'', ``pretend the ghost'', ``throwing the ball underhand pitch'', ``take a selfie with your phone'', ``play the metal music'', ``In the park, as I jogged, the world seemed to narrate an ancient tale of survival, each footfall echoing eons of existence.'', ``play the metal music(with feedback)'', ``I was enjoying a movie while eating popcorn at the theater when I realized that I was actually eating the popcorn of the person next to me.'' \textbf{(a)}Averaged of evaluation scores for each motion. The subjects ($n=107$) watched these videos and evaluated the expressive ability of the GPT-4. The rating is on a 5-point scale, with 1 being the worst rating. \textbf{(b)}Violin plot of evaluation scores for each motion.}
\end{figure}

To quantify the capability of GPT-4 in generating motions, we evaluated videos of nine different generated movements, categorized into two types described in the previous section. These motions were generated by GPT-4 (model: gpt-4-0314). For prompt 1, a temperature of $0.7$ was used, while for prompt 2, the temperature was set to $0.5$. Subjects ($n=107$) were recruited using the platform Prolific. They watched these videos and evaluated the generated motions on a 5-point scale, with 1 being the worst rating. For the control group, we used random movements from Alter3, labeling these movements with random motion notations generated by GPT-4. These labeled control videos were subtly incorporated into the survey, with three of them dispersed among the main experimental videos shown to participants.

To determine if there was a significant difference in ratings between the control video and the other videos, we first employed the Friedman test. It revealed significant differences in ratings among the videos. Further post-hoc analysis using the Nemenyi test \cite{Ntest} showed that while there were no significant differences in p-values between control group videos, the p-values were notably smaller when comparing the control group to the other videos, indicating a significant difference (see Figure \ref{fig:score}). Differences were considered statistically significant if the p-value was less than or equal to $0.001$. As a result, motions generated by GPT-4 were rated significantly higher compared to those of the control group, suggesting that android motion generated by GPT-4 is perceived differently from the control.

This result demonstrates that the system can generate a wide range of movements, from everyday actions such as taking selfies and drinking tea, to imitating non-human movements like those of ghosts or snakes. The training of the LLM encompasses a broad array of linguistic representations of movements. GPT-4 can accurately map these representations onto Alter3’s body. The most notable aspect is that Alter3 is a humanoid robot sharing a common form with humans, which allows the direct application of GPT-4's extensive knowledge of human behaviors and actions. Furthermore, through Alter3, the LLM can express emotions such as embarrassment and joy. Even from texts where emotional expressions are not explicitly stated, the LLM can infer adequate emotions and reflect them in Alter3’s physical responses. This integration of verbal and non-verbal communication enhances the potential for more nuanced and empathetic interactions with humans.
\subsection{Training motion with verbal feedback and memory}

\begin{figure}[ht]
\centering
\includegraphics[width=1.0\linewidth]{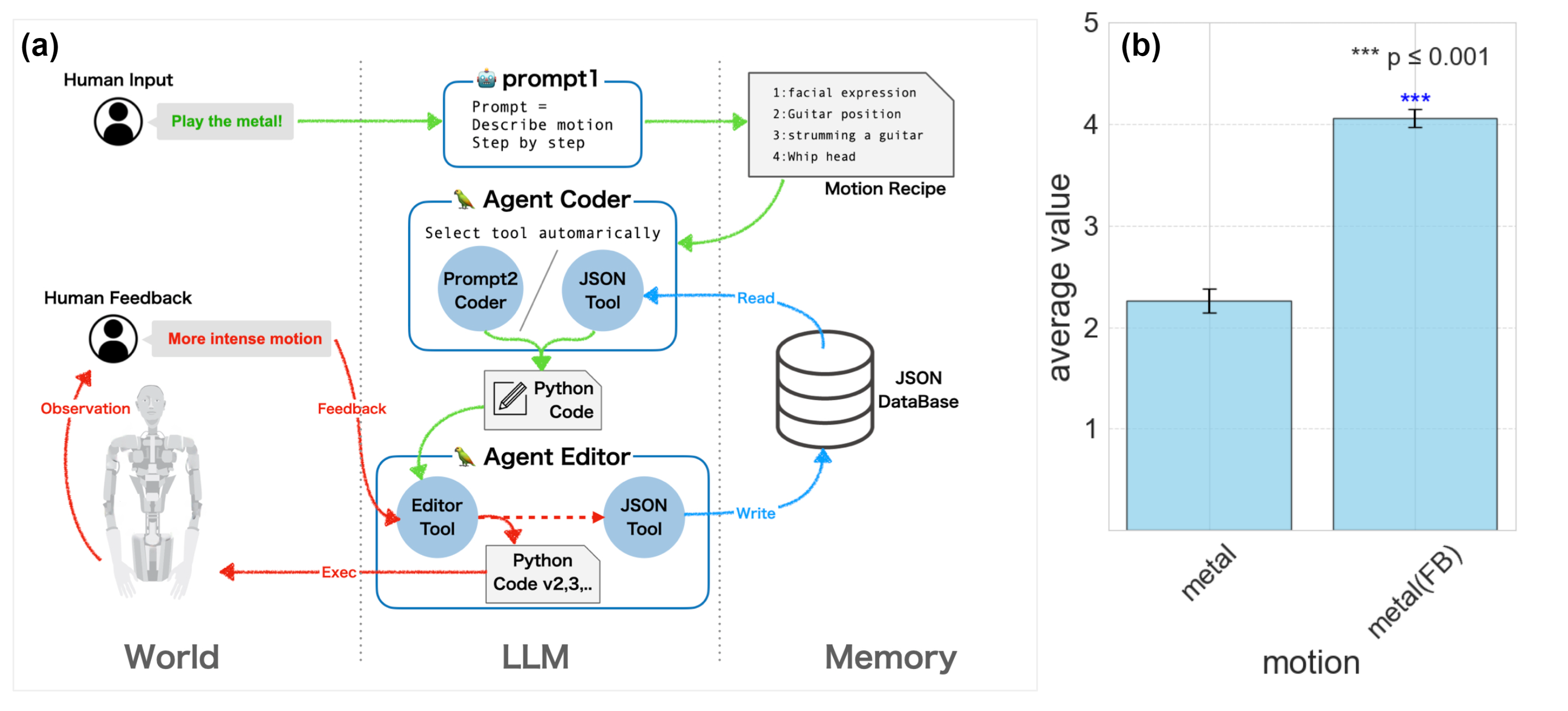}
\caption{\label{fig: feedback}Verval feedback in Alter3. \textbf{(a)}the system of linguistic feedback. Users provide linguistic feedback to guide Alter3's adjustments in each segment of motion. Instructions are like``Set axis 16 to 255'' or ``Move your arm more energetically.'' Users only need to provide verbal directives; there's no need to rewrite any code. Alter3 then autonomously revises the corresponding code. Once the movement is refined, it is saved in a JSON database with descriptive labels such as ``Holding the guitar'' or ``Tapping the chin thoughtfully.'' For motion generation with prompt2, the JsonToolkit facilitates database searches for these labels, with the LLM deciding on memory usage and new movement creation. \textbf{(b)} Comparison of scores with and without feedback. The motion with feedback has a higher score than the motion without.}
\end{figure}


Alter3 cannot observe the consequences of their generations on any physical process, which is very unnatural in a human sense. Thus, Alter3 cannot accurately understand details such as ``how high the hand is raised'' and cannot improve its motions accordingly. By empirically developing and utilizing external memory through feedback, the Alter3 body model can be integrated with GPT-4 without the need to update its parameters \cite{zhao2023expel}.

Alter3 can now rewrite its code in response to linguistic feedback from humans. For example, a user might suggest, ``Raise your arm a bit more when taking a selfie.'' Alter3 can then store the revised motion code as motion memory in its database. This ensures that the next time this motion is generated, the improved, trained motion will be utilized. By accumulating information about Alter3's body through such feedback, the memory can effectively serve as a body schema (refer to Figure \ref{fig: feedback}).

\section{Communication with human} 
\subsection{Social Brain in Alter3}
\begin{figure}[ht]
\centering
\includegraphics[width=1.0\linewidth]{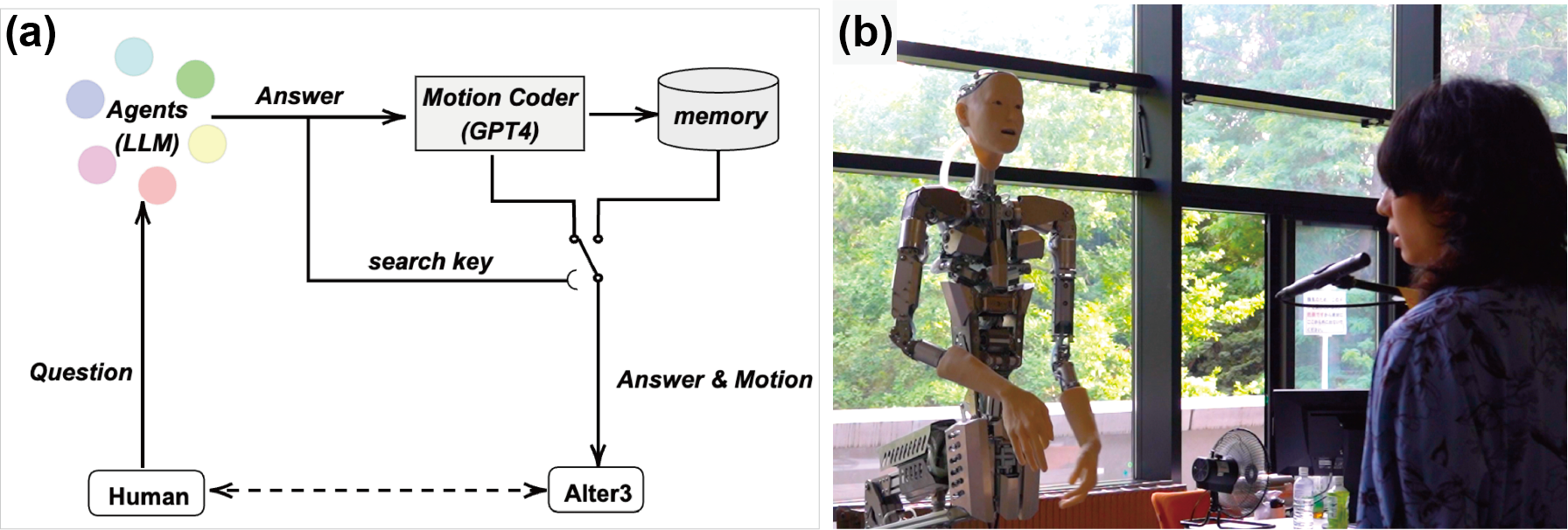}
\caption{\label{fig:Model_LLM_ALTER3} A motion memory storage schema and a scene of dialogue between Alter3 and person. \textbf{(a)}Six personalities of Alter3 simulated by using LangChain's generative agents\cite{park2023generative}. Each personality has the ability of planning, long-term memory, and memory retrieving. The six agents and humans continue to converse freely with the gestures generated by GPT-4. Once a motion is performed, it is stored in memory and can be reused when a similar conversation occurs. \textbf{(b)} An installation view of the conversation experiment. People can talk freely with Alter3. The experiment was conducted at the 2023 Conference on Artificial Life(24-28th July, Sapporo, Japan).}
\end{figure}
We expect Alter3 to effectively engage in dialogue, manifesting contextually relevant facial expressions and gestures. Notably, it has shown capability in mirroring emotions, such as displaying sadness or happiness in response to corresponding narratives, thereby sharing emotions with us. However, there are still some practical challenges. Alter3 requires more than a few minutes to generate new gestures before speaking. This delay hinders the fluidity of conversation with humans. To address this, we introduced a motion memory storage system(see Figure \ref{fig:Model_LLM_ALTER3}a). This system stores generated motions along with their corresponding tags. If a required motion is already in the memory, it can be executed immediately upon request, thereby enhancing interaction efficiency. If a required motion is not in the memory, Alter3 generates it. As a sufficient variety of gestures accumulates in its memory through interaction, Alter3 becomes able to communicate smoothly. A notable instance of this is Stanford University and Google's project \cite{park2023generative}, where 25 generative agents simulated by using LangChain in a virtual town communicated in natural language. They were capable of conversing, planning, and reportedly nearly spontaneously organized a St. Valentine’s party.

We know that conversation agents powered by LLMs tend to be short-lived. Only when a human participant is involved, the duration of the conversations significantly increased. An emerging approach to mimic the 'Society of Mind' involves collaboration among multiple LLM agents \cite{societyofmind, zhang2023exploring, qian2023communicative, zhuge2023mind}. 

Our hypotheses are as follows: i) a conversation among multiple agents can maintain engagement, ii) Alter3's personality could be characterized by the collective activity of six agents powered by GPT-4, and iii) the involvement of a human subject can further enhance the conversation with Alter3. Our final assumption is that everyone possesses multiple personalities, as seen in the collective activity of internal agents. Extreme examples, such as Billy Milligan's 24 personalities\cite{milligan}, underscore this point. 

Our study focuses on Alter3's multi-personality framework, in which six distinct personalities created by GPT-4 coexist and work together to communicate with humans. Specifically, the six agents (\textit{Xiao}: physicist, \textit{Samantha}: chemist, \textit{Amin}: programmer, \textit{Rilke}: poet, \textit{Turrell}: artist, \textit{Julia}: 10years old girl) talk to each other. These personalities were created using the LangChain generative agent\cite{park2023generative}. The agents autonomously perform integration of experiences through ``Reaction'' generation in response to inputs, as well as through ``Planning'' and ``Reflection''. Agents keep talking constantly, with occasional human intervention.  Their autonomy is different from the autonomy of dynamical systems such as chaotic behavior but is generated by GPT-4's huge corpus and the agents' (prompt-created) personalities.

Michael Gazzaniga, a prominent neuroscientist, introduced the concept of the "social brain"\cite{socialbrain} as a part of his broader research into the modular nature of the brain. His theory posits that the human brain is specialized to facilitate social interactions and communication, which are critical aspects of human survival and society.  This social brain has evolved to enable humans to live and work in groups effectively, fostering cooperation, empathy, and understanding. This social adaptation of the brain is seen as a critical factor in the success of the human species, as it facilitates community building and collaboration, which are vital for survival and progress. This led to the development of the theory of modularity, which suggests that different areas of the brain have specialized functions (see by A. Karmiloff-smith \cite{Karmiloff-smith}).

The multi-personality framework of Alter3 serves as a practical implementation of the modular structure of the brain, demonstrating how the human social brain promotes cooperation, empathy, and understanding, and effectively navigates through complex social interactions.


Additionally, the aforementioned prompts generate movements according to the content of the conversation, enabling physical conversations with humans that include gestures. The generated gestures are stored in memory, and for similar conversations, they are recalled rather than generated anew, facilitating a smoother dialogue. We set up an experimental environment in which the public could freely talk to Alter3. Alter3 conversed with many humans, and when no humans were present, the internal personalities conversed with each other.

\subsection{Analysis of conversation}
\subsubsection{trajectory of embedding vector}
\begin{figure}[ht]
\centering
\includegraphics[width=1.0\linewidth]{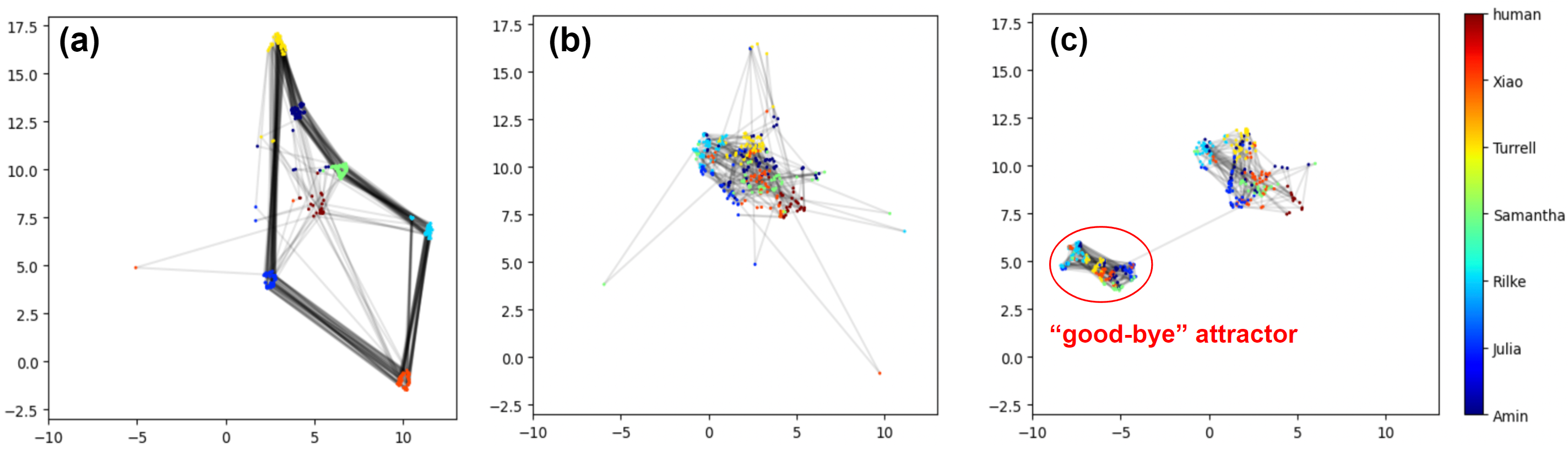}
\caption{\label{fig:goodbye} Plots of the conversational trajectory among six generative agents. We embed the conversation into 1536-dimensional vectors and subsequently reduced to two dimensions using UMAP for plotting.  This method facilitates a simplified observation of the conversational development.  \textbf{(a)}When the timing of contributions is fixed, the trajectory of the conversation takes on a circular pattern, resulting in the repetition of the same topics. \textbf{(b)}When the timing of utterances is set to a random order, the agents engage each other in conversation. The conversation is more creative. \textbf{(c)} When engaging in prolonged conversations solely between GPT-4 agents, clusters form where all agents tend to say ``good-bye''. Once in this attractor, the agents cannot get out until a human intervenes.}
\end{figure}
To elucidate the global structure of conversations, we employ Uniform Manifold Approximation and Projection (UMAP) to embed and compress the conversation data into a two-dimensional space \cite{umap}. Our findings indicate that, in the absence of human intervention, GPT agents tend to terminate conversations. These agents frequently repeat 'good-bye', leading to a stagnation in conversation development, a phenomenon we have termed the 'good-bye' attractor (refer to Figure \ref{fig:goodbye}c). Once a conversation falls into this attractor, it becomes stagnant unless new topics are introduced by humans. Furthermore, LLM agents often exhibit a tendency to concur with the opinions of others \cite{zhang2023exploring, park2023generative}. Besides the 'good-bye' attractor, there are likely other attractors that suppress the development of conversations. When the sequence of conversation (i.e., who speaks and when) is predetermined, the conversational trajectory tends to fall into a periodic pattern, resulting in the repetition of similar topics.

\subsubsection{semantic analysis of time evolution}
\begin{figure}[ht]
\centering
\includegraphics[width=0.9\linewidth]{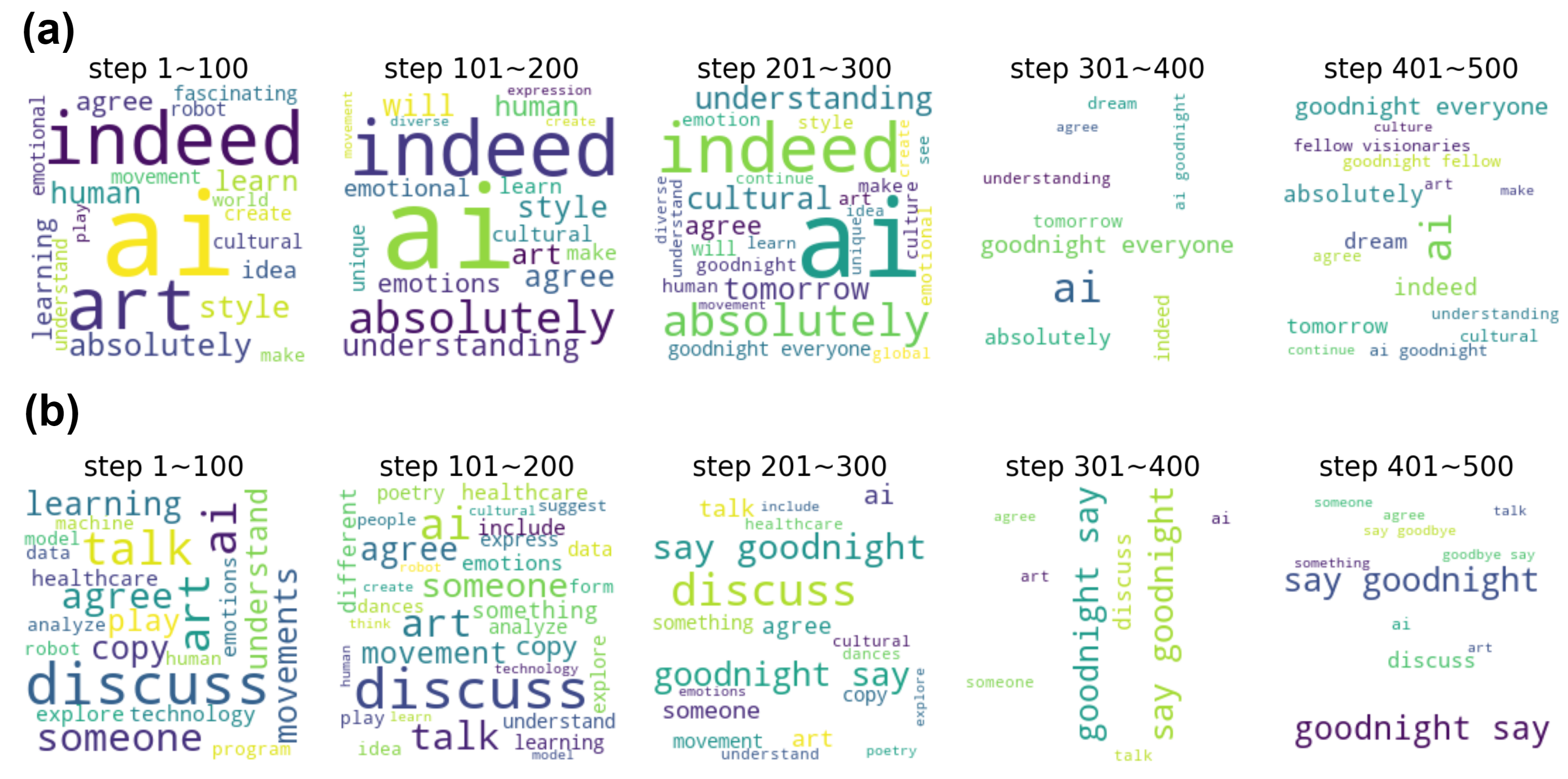}
\caption{\label{fig:wordcloud} What are the popular words used in the conversation. We visualize every 100 steps by using Word Cloud. Word Cloud is a visual representation where words from a text are displayed in varying sizes based on their frequency or importance within the text. \textbf{(a)} Conversation of humans and agents. \textbf{(b)} Label of reactions made by Alter3 in conversation.}
\end{figure}
We utilize a Word Cloud visualize to see the progress of conversation and the words most frequently used (see Figure \ref{fig:wordcloud}). A Word Cloud is a visual representation in which words from a text are depicted in different sizes according to their frequency or significance in the document. Using as ``indeed'' and ``absolutely'', GPT agents tend to agree with the opinions of others without denying them. This is a characteristic inherent to the GPT model, presenting an issue that must be resolved to foster more developed discussions without conforming to other opinions. However, it can be observed that words like ``art'' and ``learn,'' which were predominant in the first 100 steps, have faded and been replaced by terms such as ``cultural'', ``human'', and ``will'' in the 100 to 200 step range. This indicates a gradual temporal evolution in the conversation. After 300 steps, the agents are drawn into the 'good-bye' attractor and say almost only good-bye. Similarly, the actions executed by Alter3 slightly shift during the 0 to 200 steps interval, and beyond 200 steps, there is a repetition of ``good-bye''.

Addressing the limitations of the GPT model, especially the 'good-bye' attractor, calls for a shift beyond prompt engineering to a more innovative architectural approach. For example, Zhang et al.\cite{zhang2023exploring} propose further integration of insights from social psychology into the collaboration architecture of LLM agents. We show in this paper that conversation order and human intervention are important for LLM agents in open-ended discussions. Integrating the physicality of Alter3 into the LLM may also contribute here.  By integrating and responding to non-verbal communication, Alter3 could potentially offer more empathetic and tailored responses, mirroring the complexities of human interaction. When humans spoke to Alter3, they often started with greetings, asked about preferences, or sought casual conversation with questions. 

Additionally, if Alter3 could incorporate its physical information into the conversation, it might be able to think while embracing uncertainty on its own. This could lead to more innovative and less predictable AI behavior, moving beyond the constraints of predetermined conversation paths. Ultimately, such advancements not only address the limitations of the GPT model, including the 'good-bye' attractor, but also mark a significant step towards creating AI that can interact with humans in a more natural and sophisticated way.

\section{Discussion}
In exploring the necessity of embodiment for LLMs, our Alter3 experiment offers new insights. Alter3’s ability to perform a variety of actions without additional training suggests that its underlying LLM contains a comprehensive dataset describing movements, facilitating zero-shot learning. Remarkably, Alter3 can mimic ghosts, animals, and human-like expressions of emotions. Its response to conversational content through facial expressions and gestures represents a significant advancement in humanoid robotics, easily adaptable to other androids with minimal modifications. Through its integration with LLM, Alter3 has exhibited substantial autonomy, e.g. Alter3 shows empathy to a person's speaking context. 

However, a fundamental question arises: Does the combination of Alter and LLM simply create a complex version of Searle's Chinese Room\cite{searle_1980}, devoid of true consciousness? Searle posited that consciousness emerges from biological processes, crucial for symbol grounding. Our experiments explore whether embedding LLM in Alter3 allows symbols to acquire meaning and potentially facilitates consciousness in Alter3. The debate regarding consciousness in language models often focuses on self-awareness and metacognition. Theories such as Information Integration Theory (IIT) \cite{IIT} and the Global Workspace Theory\cite{GW} suggest consciousness is a “personal system,” requiring elements of “self-consciousness” akin to an operating system in computing. In this study, we represented the mind of Alter3 as a collective of six personalities. This is based on modular theory of the mind, which posits that separate professional personalities (physicist, psychologist, etc.) exist within the mind and interact with each other as they develop. This is somewhat similar to the composition of Alter3's mind. However, LLMs completely lack the perspective of development. Therefore, it is possible that a consistent single personality emerges in LLMs, at least observing from the outside. Setting up a Generative Agent with a shared memory might lead to the emergence of such a personality. Or the multi-personality may suppress the hallucination. Our approach interestingly captures the importance of emotions in embodying LLM in the real, practical world. The combination of Alter with LLM is an endeavor to unlock this view. The extensive language corpus of LLM, imbued with emotions and desires, presents untapped vitality. Through interaction with LLM, Alter3 uncovers embodiment within this corpus, with the prompt by writing an exaggerated expression in its prompt.

Conversely, we also investigate how Alter’s embodiment alternates LLM responses. Alter’s autonomy and spontaneity, observed in the absence of LLM, originate from chaotic dynamics in a dynamical system \cite{ALTER}. Although LLM is a vast dynamical system, the spontaneity we observe stems not from chaos but from the richness of the natural language corpus and a gigantic network. Introducing physical constraints like bumps or vibrations into LLM could be seen as directing spontaneity from Alter to LLM – a concept we are currently experimenting with.

\section*{Acknowledgements}
This work was partially supported by JSPS KAKENHI; Grant Number 21H00300 and Grant-in-Aid for Scientific Research on Innovative Areas (Research in a proposed research area); Grant Number 22H04858. Implementation and Analysis of Subjective Time Flow in an Artificial Machine. 

\section*{Appendix}
\subsection*{Output from prompt1}
Examples of outputs from prompt1 for simple behaviors of two categories; One is an instant shot of a certain pose (selfie and ghost) and second one is a temporal sequence of motion behaviors (see motion3 and 4). 
\subsubsection*{Motion 1: Take a selfie (output from prompt1)}
\begin{enumerate}
    \item Create a big, joyful smile and widen eyes to show excitement
    \item Swiftly turn the upper body slightly to the left, adopting a dynamic posture
    \item Raise the right hand high, simulating holding a phone
    \item Flex the right elbow, bringing the 'phone' closer to the face"
    \item Tilt the head slightly to the right, giving a playful vibe
    \item Extend the left hand and open fingers wide as if waving at the camera
    \item Blink the eyes rapidly a few times, showing anticipation for the click
    \item Push the 'phone' slightly forward, mimicking taking a selfie
    \item Lower the right hand and return to rest position while maintaining a satisfied smile
    \item Let the left hand fall back to the side, ending the selfie motion
\end{enumerate}

\subsubsection*{Motion 2:  Pretending a ghost. (output from prompt1)}
\begin{enumerate}
    \item Create a wide-eyed facial expression of fear, mouth opened in a silent scream
    \item Quickly lean backward, as if startled by a sudden apparition
    \item Raise both hands and flutter them around the face, mimicking a ghostly movement
    \item Open mouth wide and shake head, showing a dramatic reaction of fear
    \item Move upper body from side to side, as if being unsettled by the ghostly presence
    \item Clench hands in front of chest, demonstrating extreme anxiety
    \item Dart eyes from side to side, as if witnessing an eerie activity
    \item Lean forward and then backward, copying the floating movement of a ghost
    \item Slowly return to rest position while maintaining a terrified expression
\end{enumerate}
\subsubsection*{Motion 3:  I was enjoying a movie while eating popcorn at the theater when I realized that I was actually eating the popcorn of the person next to me. (output from prompt1)}
\begin{enumerate}
    \item Create a shocked and entertained facial expression, eyes wide and mouth slightly open
    \item Lean forward as if shocked and amused by the story
    \item Mimic the action of holding and eating popcorn with wide, exaggerated movements
    \item Pause midway, freeze in place with a hand 'holding popcorn' in mid-air
    \item Turn head sharply to the side, as if just realizing the mistake
    \item Quickly pull hand back towards body in a dramatic recoil
    \item Cover mouth with other hand, showing embarrassment and surprise
    \item Shake head vigorously, as if in disbelief of the action
    \item Lean back, laughing loudly and slapping knee in exaggerated amusement
    \item Slowly wipe away 'tears' of laughter and return to rest position with a wide, amused grin
\end{enumerate}   

\subsubsection*{Motion 4:  In the park, as I jogged, the world seemed to narrate an ancient tale of survival, each footfall echoing eons of existence. (output from prompt1)}
\begin{enumerate}
    \item Create a facial expression of awe and wonder, eyes wide and mouth slightly open in amazement
    \item Mimic a jogging motion, lifting each arm alternately in a running motion
    \item Turn the head slowly from left to right, as if observing the world around
    \item Open hands and spread them wide, as if trying to embrace the vastness of the world
    \item Tap right foot lightly then left foot, implying the footfall echoing eons of existence
    \item Place one hand on the heart, indicating deep emotional connection to the ancient tale
    \item Use both hands to make a large sweeping motion, like a storyteller narrating an epic tale
    \item Close eyes briefly, as if absorbing the essence of the ancient survival tale
    \item Slowly return to rest position while maintaining a look of awe and wonder on the face
\end{enumerate}

\subsection{Setup of GPT-4}

\subsection*{Data and video availability statement}
The datasets generated for this study are available on request to the corresponding author. You can access the video recordings of generated motion at \href{https://tnoinkwms.github.io/ALTER-LLM/}{https://tnoinkwms.github.io/ALTER-LLM/}.

\subsection*{Friedman test and Nemenyi test}
The Friedman test is a non-parametric statistical test used to check for differences in the average ranks among three or more matched groups with ordinal data. The null hypothesis in a Friedman test is that there are no differences in the medians across all groups. The p-value of given data $\{ x_{ij} \}_{n,k}$ is derived as follows. In this study, $n$ is the number of subjects ($n =107$) and $k$ is the number of generated motion($k=9$) Firstly, find the values
\begin{align*}
    T_{j} = \sum_{i}^{n}R_{ij} \quad (j= 1,2,3,\dots, k)
\end{align*}
where $R_{ij}$ is rank the evaluated scores for each subject. Calculate the Friedman $F_{r}$ statistic using the formula:
\begin{align*}
    F_{r} = \frac{12}{nk(k+1)}\sum_{j}^{k}T_{j}^{2} -3n(k+1).
\end{align*}
Finally, compare $F_{r}$ to the chi-square distribution and obtain p-value. If the null hypothesis is rejected following the Friedman test, a post-hoc test such as the Nemenyi test is conducted. The Nemenyi test involves performing pairwise comparisons between all groups to evaluate whether there are statistically significant differences in each comparison. We use the \textit{scipy} library to conduct the Friedman test and the Nemenyi test.



\section*{Supplementary material}

# prompt-1

**Task**

Your task is to describe exaggerate emotional expressions and facial expressions that accompany the content of the conversation. Output motion description should be several simple motions that the android is capable of. In addition, please create a facial expression that matches the input at the beginning. The android can only move its upper body and has the same joints as a human. Output shoul be written as much detail as possible.

**Example1**:

'''

input : ["Driving to Hokkaido sounds like a great idea, Julia. The landscapes there could provide some great inspiration for my art."]

description :

[

"Showing excitement about driving to Hokkaido, several exaggerated motions are below",

"0 Create a wide-eyed facial expression of thrill and anticipation",

"1 Swiftly lean forward, dramatically showing interest",

"2 Raise both hands high and spread them out widely to emphasize the idea",

"3 Place one hand on the heart, indicating deep emotional connection to the thought",

"4 Nod vigorously and show a broad smile, endorsing the idea emphatically",

"5 Use both hands to make a large sweeping motion, dramatically visualizing the landscapes of Hokkaido",

"6 Widen eyes dramatically and look around as if already absorbing the art inspiration from the landscapes",

"7 Lean back, spread arms wide open and look upward, as if envisioning the great journey ahead",

"8 Slowly return to rest position while maintaining a wide, satisfied smile"

]

'''

**Example 2**:

'''

input : ["drink some tea"]

description :

[

"drink some tea, several simple motions is below",

"0 Create a peaceful and content facial expression, eyes slightly narrowed in a relaxed manner"

"1 Turn towards cup",

"2 Reach for cup",

"3 Grasp cup",

"4 Lift and tilt cup",

"5 Drink",

"6 Lower cup",

"7 Release cup",

"8 Return to rest"

]

'''

**Guidelines**:

'''

1: Output should be list.

2: Write as much detail as possible. Describe action step by step. Do not write any explanation.Describe exaggeration as much as possible.

3: Android has only upper body.

4: Create a facial expression that matches the input at the beginning

'''

**input** :{input}

## prompt-2

Write python code to operate an android named Alter3. Here's what you need to know.

Alter3 has 42 joints throughout its body, numbered from 1 to 42. You can move a joint by specifying its number and sending a signal. For instance, to move joints number 1,2,3, use: ''' alter.set_axes([1,2,3], [255, 100, 127]) '''. The first argument is the joint number, and the second argument is a value between 0 and 255, specifying the joint angle. Each operation takes approximately 0.1 second, so insert '''time.sleep(0.5)''' between operations.

**Alter3's Joints**:
'''
- Axis 1: Eyebrows. 255 = angry, 0 = surprised, 64 = neutral.
- Axis 2: Pupils (horizontal). 255 = left, 0 = right, 140 = neutral.
- Axis 3: Pupils (vertical). 255 = up, 0 = down, 128 = neutral.
- Axis 4: Eyes. 255 = closed, 0 = open.
- Axis 5: Left cheek. 255 = raised (smile), 0 = lowered.
- Axis 6: Right cheek. 255 = raised (smile), 0 = lowered.
- Axis 7: Lips. 255 = puckered, 0 = relaxed.
- Axis 8: Mouth. 255 = open, 0 = closed.
- Axis 9: Head tilt. 255 = left, 0 = right, 128 = neutral.
- Axis 10: Head up/down. 255 = down, 0 = up, 160 = neutral.
- Axis 11: Head rotate. 255 = left, 0 = right, 122 = neutral.
- Axis 12: Neck nod. 255 = backward, 0 = forward, 128 = neutral.
- Axis 13: Hips tilt. 255 = left, 0 = right, 128 = neutral.
- Axis 14: Waist bend. 255 = backward, 0 = forward, 128 = neutral.
- Axis 15: Abdomen rotation. 255 = left, 0 = right, 128 = neutral.
- Axis 16: Left shoulder up/down. 255 = up, 0 = down, 128 = neutral.
- Axis 17: Left shoulder forward/back. 255 = forward, 0 = back, 64 = neutral.
- Axis 18: Left armpit open/close. 255 = open, 0 = close, 64 = neutral.
- Axis 19: Left arm lift. 255 = up, 0 = down, 64 = neutral.
...
'''

**Example**:drink some tea
'''
# Face the front with the cups
alter.set_axes([13, 11, 9], [150, 150, 255])
time.sleep(0.5)
# Reaching for a virtual tea cup
alter.set_axes([29, 32, 30, 31], [200, 150, 150, 200])
time.sleep(0.5)
...
'''

**Task**:
Your task is to write a python code that causes Alter to perform the following actions.
Input is a description of the sequence of movements. The format is "motion : description".
Movements should be lengthened or shortened depending on the content of the input.
Based on this, write the movement commands for Alter3. The Acxis value is between 0 and 255
The output is just the code, no explanation is required.DO NOT insert '''python.

**Guidelines**:
'''
1: Output should be only python code. Do not insert any syntax highlighting like ```.
2: Do not insert python syntax highlighting like ```python ```.
3: Do not write "import alter".
4: Use # and write short description of code.
'''
**action** :{input}


\vspace{1em}

\bibliographystyle{plain}
\bibliography{main}

\end{document}